\documentclass[twocolumn]{article}
\usepackage{preprint}
\usepackage{framed,multirow}

\usepackage{amssymb}
\usepackage{latexsym}
\usepackage{cite}
\usepackage{amsmath,amssymb,amsfonts}
\usepackage{algorithmic}
\usepackage{graphicx}
\usepackage{textcomp}
\usepackage{multirow}
\usepackage{subcaption}
\usepackage[colorlinks]{hyperref}
\usepackage{xcolor}
\usepackage{pifont}

\usepackage{url}
\usepackage{xcolor}
\definecolor{newcolor}{rgb}{.8,.349,.1}

\title{Fractional Concepts in Neural Networks: Enhancing Activation Functions}
\author{
    \href{https://orcid.org/0000-0002-1448-9068}{\includegraphics[scale=0.06]{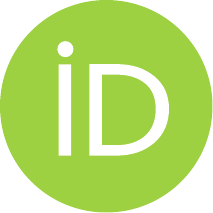}\hspace{1mm}Zahra Alijani} \\
	Institute for Research and Applications of Fuzzy Modeling \\
    University of Ostrava \\
    30. dubna 22, 701 03 Ostrava \\
	\texttt{zahra.alijani@osu.cz} \\   
	\And
	\href{https://orcid.org/0000-0002-0499-0481}{\includegraphics[scale=0.06]{orcid.pdf}\hspace{1mm}Vojtech Molek} \\
	Institute for Research and Applications of Fuzzy Modeling \\
    University of Ostrava \\
    30. dubna 22, 701 03 Ostrava \\
	\texttt{vojtech.molek@osu.cz}
}

\begin{document}
\twocolumn[\begin{@twocolumnfalse}
\maketitle

\begin{abstract}
Designing effective neural networks requires tuning architectural elements. This study integrates fractional calculus into neural networks by introducing fractional order derivatives (FDO) as tunable parameters in activation functions, allowing diverse activation functions by adjusting the FDO.
We evaluate these fractional activation functions on various datasets and network architectures, comparing their performance with traditional and new activation functions. Our experiments assess their impact on accuracy, time complexity, computational overhead, and memory usage.
Results suggest fractional activation functions, particularly fractional Sigmoid, offer benefits in some scenarios. Challenges related to consistency and efficiency remain. Practical implications and limitations are discussed.
\end{abstract}
\end{@twocolumnfalse}]

   \label{sec:intro}   
    \begin{figure}
        \centering
        \includegraphics[width=\linewidth]{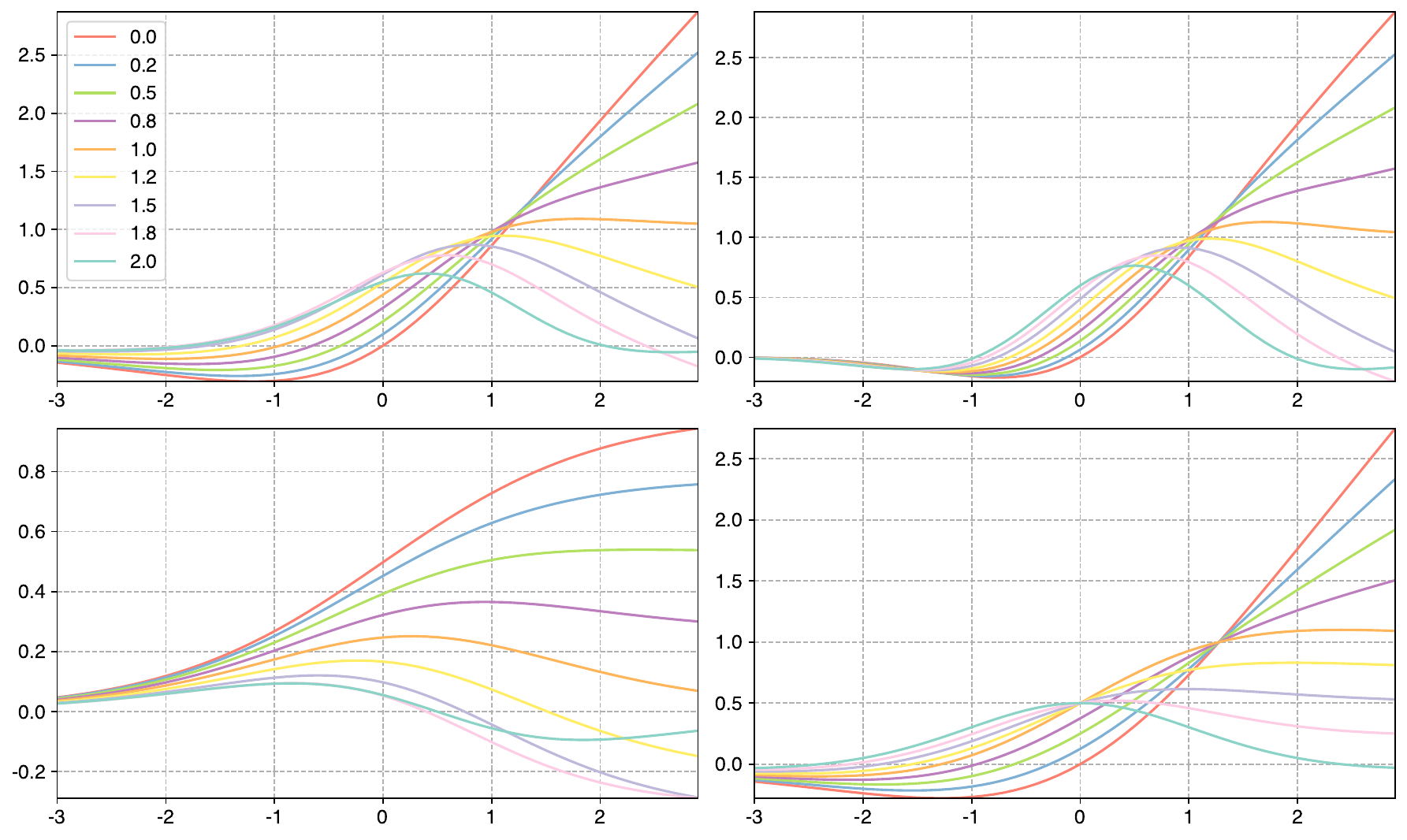}
        \caption{Fractional activation functions visualized. The functions in a row-major order are fractional Mish, fractional GELU, fractional sigmoid, and FALU~\cite{zamora2022fractional}. Graph lines represent original functions (0.0 line) and their fractional derivations.}
        \label{fig:frac_act_fcs}
    \end{figure}

    Fractional calculus pushes the boundaries of traditional derivatives and integrals by embracing orders that deviate from integer values~\cite{ortigueira2004differences}. The unique properties of fractional calculus have been shown to enhance the performance of artificial neural networks (ANNs) across various tasks~\cite{aguilar2020fractional, wu2017mittag}.

    In this paper, we revisit and expand upon the work of~\cite{zamora2019adaptive, zamora2022fractional}. The activation function is a fundamental component within a neural network, introducing the essential nonlinearity required to model complex relationships in machine learning tasks, including classification and regression. This paper emphasizes the critical importance of selecting appropriate activation functions. Over the years, various neural network architectures have emerged, each accompanied by its distinctive set of activation functions—ranging from sigmoid, radial basis function~\cite{broomhead1988radial}, ReLU~\cite{fukushima1969visual}, Softplus~\cite{dugas2000incorporating}, Swish~\cite{ramachandran2017searching}, Mish~\cite{misra2019mish}, and others. However, current practice predominantly relies on manual or automated~\cite{elsken2019neural} selection of activation functions, often leading to an exhaustive trial-and-error methodology and frequent retraining to find the optimal configuration.

    Fractional calculus opens the door to entire families of functions by naturally extending the original function through fractional derivatives. The user only needs to specify the original function, while the fractional derivative is automatically adjusted. Parameterization with a trainable argument places fractional activation functions in the same category as adaptive activation functions like PReLU or Swish~\cite{he2015delving}. However, the application of fractional activation functions comes with its own set of challenges. One prominent challenge is the increased computational and memory complexity associated with fractional derivatives, which we discuss in Section~\ref{sec:hp_tuning}. Moreover, research on fractional activation functions is limited, leaving the practical effectiveness of these functions still under-explored.

    The investigation of fractional activation functions for neural networks is a developing domain. In~\cite{ivanov2018fractional}, the authors introduce Mittag-Leffler (M-L) functions, a class of parametric transcendental functions that generalize the exponential function. They present the derivative formula for M-L functions, showing its relation to the exponential function. Activation functions in ANNs are then discussed, with popular choices including sigmoidal functions such as the logistic function. Some parameter adjustments led to improved accuracy in tests for approximating the logical OR operation and in a classification task using a simple multi-layer perceptron (MLP) with two hidden neurons. Although the results do not offer a universal parameter selection rule, empirically effective settings can be identified with minimal effort. Fractional order values near 0 or 2 generally performed better. Increasing M-L calculation precision initially boosted accuracy, but performance plateaued beyond 260 terms.

    In~\cite{zamora2019adaptive}, the authors introduced a selection of fractional activation functions derived from Softplus, such as ReLU, hyperbolic tangent, and sigmoid. A combination of ResNet-18/20 and fractional activation functions outperformed ResNet-100/110 and standard activation functions. The networks were trained and tested on CIFAR-10~\cite{krizhevsky2009learning} and ImageNet-1K~\cite{ILSVRC15}. However, the authors directly compare their fractional activation function results with those of~\cite{he2016deep} without providing an explanation of their training routine or a public repository.

    The study in~\cite{zamora2022fractional} presents the fractional adaptive linear unit (FALU), an activation function that generalizes earlier fractional models discussed in~\cite{zamora2019adaptive}. The authors manipulated and simplified formulas to derive final equations that eliminate computationally difficult terms such as the gamma function $\Gamma$. However, this simplification is not universally applicable to all fractional activation functions. Once again, the authors compared the results of established and custom ResNet architectures without explaining their training routine or providing a public repository.

    The paper by~\cite{job2022fractional} introduced fractional ReLU (FReLU) and its variations using the expansion of the Maclaurin series. It highlights the flexibility of these functions compared to standard activation functions. Regression simulation studies were conducted to predict wind power generation, demonstrating the performance of ANNs (simple MLPs with one or two hidden layers) with fractional activation functions. In particular, fractional PReLU (FPReLU) and fractional LReLU (FLReLU) consistently outperformed their standard counterparts, with FLReLU exhibiting superior performance over LReLU.

    To the best of our knowledge, the most recent effort in this area, published in 2024~\cite{kumar2024enhancing}, introduces activation functions derived using the improved Riemann–Liouville conformable fractional derivative (RLCFD). In their experiments, the authors used MLPs with one or two hidden layers and trained on simple datasets such as the IRIS~\cite{misc_iris_53}, MNIST~\cite{deng2012mnist}, and FMNIST~\cite{xiao2017fashion} datasets, in some cases achieving 100\% test accuracy. However, in these experiments, the fractional order was a non-trainable parameter and had to be chosen manually, which significantly reduced the function's potential.

    In general, to the best of our knowledge, there is no publicly available implementation of any fractional activation functions. All the above-mentioned papers, except for~\cite{zamora2019adaptive, zamora2022fractional}, study fractional activation functions on limited datasets and simple shallow MLPs.

    \textbf{The main goals} of our contribution are to evaluate published and new fractional activation functions using standard ANNs and datasets. The second goal is to provide a functional repository\footnote{\url{https://gitlab.com/irafm-ai/frac_calc_ann}} with reproducible experiment settings.

\section{Fractional Calculus}
\label{sec:frac_calculus}

    Fractional calculus is an emerging area of research, with its application in neural networks still being experimental. Calculating the fractional derivative of certain activation functions is challenging, as these may not have simple closed-form expressions. In such cases, approximations like the Grünwald-Letnikov method~\cite{ortigueira2004differences} or other specialized techniques are generally used. These methods involve approximating the fractional derivative using finite differences or numerical integration. Fractional derivatives and integrals are utilized in neural networks to modify the shape of activation functions during training. This modification is achieved by adjusting the FDO(s), which are trainable numerical parameter(s).

    \subsection{Fractional Derivatives}
        \label{subsec:frac_derivatives_and_integrals}

        Fractional calculus is a powerful mathematical tool for modeling various complex engineering and real-world systems. Three popular fundamental definitions of fractional derivatives are~\cite{herrmann2011fractional}:

        \textbf{The Grünwald-Letnikov}~\cite{ortigueira2004differences} derivative of fractional order \( a \in \mathcal{R}^+ \) is defined as:

            \begin{equation}
                \label{eq:gru_let}
                D^a f(x) = \lim _{h \to 0} \frac{1}{h^a} \sum_{n=0}^{[\frac{x-a}{h}]}(-1)^n C_{n,a}f(x-nh),
            \end{equation}
     
        where \( [x] \) denotes the integer part of \( x \) and \( C_{n,a} \) is the binomial coefficient. 
        Grünwald-Letnikov fractional derivatives are often preferred in neural networks because they are easy to compute numerically and can be applied to various activation functions.

        This expression involves an infinite sum, making it challenging to represent directly as a convex combination. However, we can approximate it by considering a finite number of terms in the sum. Let's denote this finite approximation as \( F_k(x) \), where \( k \) is the number of terms considered in the sum. Then, we have:
      
            \begin{equation}
                F_k(x) = \frac{1}{h^a} \sum_{n=0}^{k} (-1)^n \frac{\Gamma(a + 1)}{\Gamma(n + 1)\Gamma(1 - n + a)} \cdot f(x - nh).
            \end{equation}
       
        Now, \( F_k(x) \) can be considered as a convex combination of \( f(x - nh) \) for \( n = 0, 1, \dots, k \) with appropriate weights determined by the coefficients of the sum. Therefore, \( F_k(x) \) can be expressed as:
        \[
        F_k(x) = \sum_{n=0}^{k} w_n \cdot f(x - nh),
        \]
        where \( w_n \) are the weights associated with each term in the sum. This representation demonstrates that \( F_k(x) \) can be seen as a convex combination of the arbitrary activation function applied to different arguments \( x - nh \).

        \textbf{The Riemann–Liouville} derivative of fractional order \( a \in \mathcal{R}^+ \) is defined as:
        
            \begin{equation}
                D^a f(x) = \frac{1}{\Gamma (n-a)} \frac{d^n}{dt^n}\int_{a}^x \frac{f(\mu)}{(x-\mu)^{a-n+1}}d\mu,
            \end{equation}
      
        for \( n - 1 < a < n \), \( n \in \mathcal{Z}^+ \), and \( \Gamma (.) \) is the Gamma function (we will recall it shortly).
        The Riemann-Liouville derivative is used when the initial conditions are given in terms of Riemann-Liouville fractional integrals.

        \textbf{The Caputo} derivative of fractional order \( a \in \mathcal{R}^+ \) is defined as:
            \begin{equation}
                D^a f(x) = \frac{1}{\Gamma (n-a)}\int_{a}^x \frac{f^n(\mu)}{(x-\mu)^{a-n+1}}d\mu,
            \end{equation}
        for \( n - 1 < a < n \), \( n \in \mathcal{Z}^+ \), where \( f^n(\mu) \) is the \( n \)-th order derivative of the function \( f(x) \).
        The Caputo fractional derivative is often used when the initial conditions are specified as conventional (integer order) derivatives.

        The Gamma function, denoted by \( \Gamma(z) \), is a generalization of the factorial operator and is used to define the fractional derivative in fractional calculus. The Gamma function is defined as~\cite{abramowitz1972handbook}:
        \begin{equation}
            \Gamma (z)=\int_{0}^\infty x^{(z-1)}e^{-x}dt.
        \end{equation}
        The Gamma function is defined for non-negative integers as \( \Gamma(n)=(n - 1)! \), and for other nonnegative values of \( z \) it can be computed by~\cite{abramowitz1972handbook}:
        \begin{equation}
            \Gamma(z)=\frac{e^{-\gamma z}}{z}\prod_{k=1}^{\infty}\left((1+\frac{z}{k})^{-1}e^{\frac{z}{k}}\right),
        \end{equation}
        where \( \gamma \) is the Euler-Mascheroni constant \( (\gamma = 0.57721..) \).

        The fractional derivative, represented above, can be modified by replacing the factorial with the Gamma function. The definition provided in the following statement represents the fractional derivative of the function \( f(x) = x^k \) for \( k, x \geq 0 \):
        \begin{equation}
            \label{fractional def}
            D^a x^k=\frac{\Gamma(k+1)}{\Gamma(k+1-a)}x^{k-a}.
        \end{equation}
        The Gamma function allows for the definition of the fractional derivative for non-integer values of \( k \), while the factorial is only defined for integers. This allows for a more general and flexible formulation of the fractional derivative.

        In machine learning, fractional derivatives and integrals have been used in various ways. For example, they have been used to design activation functions, as discussed in~\cite{zamora2019adaptive}. They can also be used in the design of loss functions, where they can capture more complex patterns in the data.

\section{Exploring Fractional Variants of Activation Functions}
    \label{sec:grouping_Act_fc}
    Activation functions can be categorized into families utilizing fractional calculus. By representing the fractional derivative of an activation function, other activation functions within the same family can be mathematically derived.

    It is worth noting that not all functions are suitable to be activation functions. The step function is an example of a computationally efficient function but is a poor choice for an activation function due to discontinuity and flat derivations. The Sigmoid activation function used in deep networks can cause exploding or vanishing gradients~\cite{Basodi2020}. In some cases, replacing the activation function with its fractional derivative helps to alleviate its undesirable properties. In other cases, a replacement can produce these properties, e.g., a function created by taking a fractional derivative of ReLU is discontinuous. With the FDO $a$ approaching 1, the first derivative is going towards 0. We will use $a$ to denote FDO.
    
    \subsection{Fractional GELU (FGELU)}
        GELU activation function is commonly used in transformers~\cite{vaswani2017attention}. It was introduced in \cite{hendrycks2016gaussian}.
        \begin{equation*}
            f(x) = 0.5x \left(1 + \tanh\left(\sqrt{\frac{2}{\pi}}(x + 0.044715x^3)\right)\right).
        \end{equation*}

        Fractional GELU is defined using the Grünwald-Letnikov fractional derivative:
        \begin{equation}
            \label{eq:frac_gelu}
            \begin{split}
                D^a f(x)= \lim_{h \to 0} \frac{1}{2h^a} \sum_{n=0}^{\infty} (-1)^n \frac{\Gamma(a + 1)(x-nh)}{\Gamma(n + 1)\Gamma(1 - n + a)} \cdot \\
                (1 + \tanh\left(\sqrt{\frac{2}{\pi}} \big(({x - nh}) + 0.044715 ({x - nh})^3\big) \right).
            \end{split}
         \end{equation}

    \subsection{Fractional Mish (FMish)}
        Mish activation function (used, for example, in the detection algorithm YOLOv4~\cite{bochkovskiy2020yolov4}) is defined as:
        \begin{equation*}
            f(x) = x\cdot \tanh(\ln(1+e^x)) = x\cdot \frac{(e^x+1)^2-1}{(e^x+1)^2+1}.
        \end{equation*}
        Fractional Mish is computed as:
        \begin{equation}
         \begin{split}
             D^a f(x)= \lim_{h \to 0} \frac{1}{h^a} \sum_{n=0}^{\infty} (-1)^n &\frac{\Gamma (a+1)(x-nh)}{\Gamma(n+1)\Gamma(1-n+a)}\cdot \\
                \frac{(e^{x-nh}+1)^2-1}{(e^{x-nh}+1)^2+1}.
           \end{split}
        \end{equation}

    \subsection{Fractional Sigmoid (FSig)}
        The Sigmoid is one of the well-established functions in statistics and machine learning. Due to the previously mentioned gradient issues, it is commonly used as an activation function in the last layer (to squeeze output into $[0, 1]$) but not inside the networks. The function is defined as:
        \begin{equation*}
            f(x) = \frac{1}{1+e^{-x}}.
        \end{equation*}
        The fractional Sigmoid can be implemented using the soft plus function or by directly applying the fractional derivative to the Sigmoid function:
        \begin{equation}
               \lim_{h \to 0} \frac{1}{h^a} \sum_{n=0}^{\infty} (-1)^n \frac{\Gamma (a+1) }{\Gamma(n+1)\Gamma(1-n+a) (1+e^{-x+nh})}
            \label{eq:frac_sig}
        \end{equation}

    \subsection{Fractional Adaptive Linear Unit (FALU)}
\begin{figure}[ht]
        \centering
        \includegraphics[width=\linewidth]{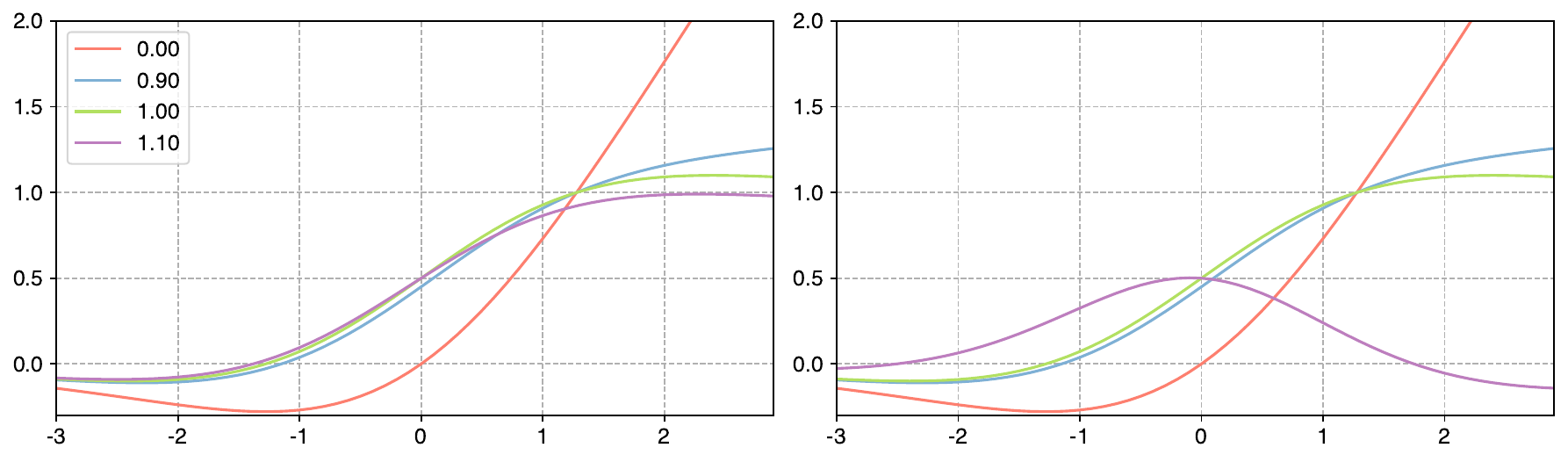}
        \caption{FALU and its fractional derivatives with fixed $\beta=1$. The left graph shows FALU with our fix, notice smooth transition between the 0.9 and 1.1 derivative. The graph on the right shows derivatives with the original FALU formulation. Notice how 1.1 derivative is approximately 2 derivative.}
        \label{fig:falu_fix}
    \end{figure}

        The FALU was introduced in~\cite{zamora2022fractional} as a flexible family of functions parameterized by two variables, $\alpha$ and $\beta$. Here $\alpha = a$ for the sake of consistency. The authors emphasize the ease of implementing this activation function in neural networks as the formula consists of simple arithmetic operations and the Sigmoid function, completely avoiding the gamma function. Although the approximation for $a \in [0, 1]$ is accurate, the approximation for $a \in (1, 2]$ is inaccurate; see Fig.~\ref{fig:falu_fix}. We propose a simple change: $a \rightarrow (a-1)$ (bold part of the formula).
        FALU limits $a \in [0, 2]$ and $\beta \in [1, 10]$ and is defined as:
        \begin{equation*} 	
        \approx
            \left\{
            \begin{array}{ll}
                g(x, \beta) + a \sigma(\beta x)(1 - g(x, \beta)), & a \in [0, 1], \\
                h(x, \beta) + \mathbf{(a-1)}\sigma(\beta x)(1 - 2h(x, \beta)), & a \in (1, 2].
            \end{array}
            \right.
      \end{equation*}
    Here, $h(x, \beta)$ is defined as $g(x, \beta) + \sigma(x)(1 - g(x, \beta))$ and $g(x, 1) = g(x) = x\sigma(x)$.

\section{Fractional Hyperparameter Tuning}
    \label{sec:hp_tuning}
    All fractional activation functions share two variables that make implementation rather difficult. The first is the variable $h$ in $\frac{1}{h^a}$ and the second is the upper bound of the summation $\infty$ in $\sum_{n=0}^{\infty}$. We will refer to the upper bound of summation as $N$. Eq.~\ref{eq:gru_let}, shows that both values are intertwined.

    As $N$ increases, so does the precision of the fractional derivative approximation, similar to the Taylor series. Examining the term $f(x-nh)$ in Eq.~\ref{eq:gru_let}, we can see that the fractional derivative at point $x$ uses the interval $[x-Nh,x]$ for computation. To keep this interval constant for different $N$, we adjust $h$:
    \begin{equation}
        \label{eq:h_rule}
        h = \frac{1}{\max(1, N-1)}.
    \end{equation}

\begin{figure}
        \centering
        \includegraphics[width=\linewidth]{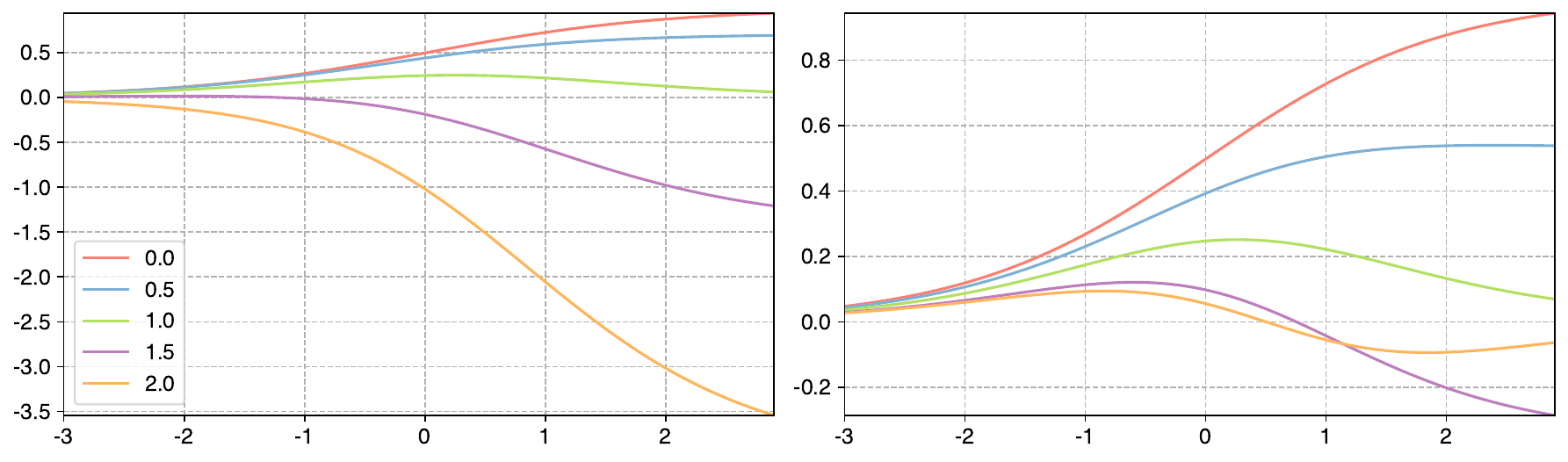}
        \caption{Fractional sigmoid with matched and mismatched $N$ $h$ pair. Left: $N=2$ and $h=0.5$. Right: $N=3$ and $h=0.5$.}
        \label{fig:h_diff}
    \end{figure}

    Fig.~\ref{fig:h_diff} illustrates the case where $h$ is set correctly and incorrectly.

    To find a suitable $N$ value, we perform training runs with varying $N$. The runs have identical training settings as described in sec.~\ref{sec:results_act_fn}, except the training dataset size and number of epochs. We train ResNet-20 on 50\% of CIFAR-10 for 100 epochs and EfficientNet-B0 on 10\% of ImageNet-1K for 30 epochs.

    \textbf{Weight decay} that we use in our experiments has a unwanted effect on the fractional order of the activation functions. It pushes the fractional orders toward zero, as the fractional orders are included in the model parameters. We prevent fractional orders decaying by setting their decay factor to $\gamma=0$.

    \begin{figure}
        \centering
        \includegraphics[width=\linewidth]{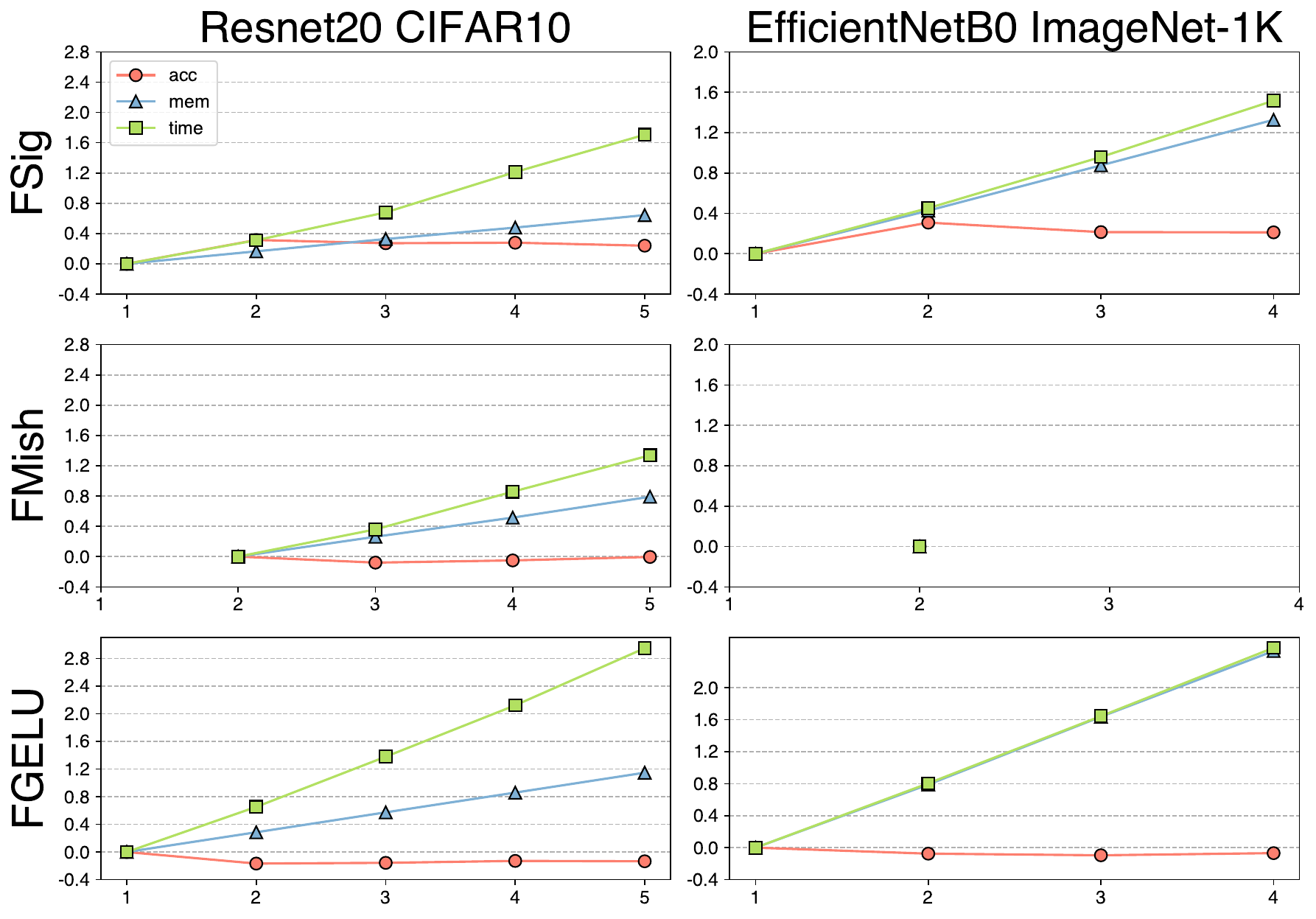}
        \caption{Test accuracies of the fractional sigmoid, Mish, and GELU in a first, second, and third row respectively. The results in the left column are obtained by training ResNet-20 on 50\% of the CIFAR-10 training set. The results in the right column are obtained by training EfficientNet-B0 on 10\% of ImageNet-1K training set.}
        \label{fig:ab_study}
    \end{figure}

    Fig.~\ref{fig:ab_study} shows $x$ increase/decrease in memory and time complexity and \textbf{test} accuracy. The missing result in the second row (FMish) is caused by the training falling into NaN. We tried to stabilize training with gradient clipping; however, the clipping norm would have to be very small and would hinder the results of other activation functions.

    In almost all runs, both the memory and the time complexity increase approximately linearly, although the time complexity increases faster. 

    The increase in memory comes from the gradient computation. The PyTorch framework, which we use for the experiments, saves immediate results during the forward pass and uses them during the backward pass. The more terms in the formula with increasing $N$, the more immediate results need to be stored.

    We used the \textbf{highest accuracy} $N$ for full-train experiments in sec.~\ref{sec:results_act_fn}. We use ImageNet-1K as a proxy for datasets with similar resolution (CalTech-256 and Food-101) to select $N$.

\section{Activation Function Performance}
    \label{sec:results_act_fn}
    We evaluate the performance of Sigmoid, Mish, and GELU and their fractional counterparts, as well as ReLU, PReLU, and FALU. We chose ReLU because it is widely used, PReLU because it includes trainable parameters, and FALU because of its promising results. All of the following results are seeded, reproducible, and use deterministic algorithms. Please refer to our repository.

    \subsection{ResNets}
    \label{subsec:resnets}
    We train several ResNet variants designed for the CIFAR-10 dataset. CIFAR-10 characteristics are the small resolution of 32$\times$32 pixels and the low number of classes (10).

    \textbf{Setting:} We train for 200 epochs with a 5-epoch warm-up. We use SGD with 0.9 momentum, 5e-4 weight decay, and an initial learning rate of 0.1 that decreases at 30\%, 60\%, and 80\% of total training steps. Data are fed to the network in batches of 128 images augmented with padded random cropping, horizontal flipping, and normalization. Furthermore, we use label smoothing 0.1, gradient clipping by max norm 10, and 16-bit floating precision. We track and report the best overall test accuracy. The fractional activation functions have $N$ and $h$ set according to sec.~\ref{sec:hp_tuning} results.

    \begin{table}
    \caption{CIFAR-10 test accuracies of different ResNet/activation function combinations.}
        {\begin{tabular}{l | c c c c c}
            \textit{ResNet} & 20 & 32 & 44 & 56 & 110 \\
            \textit{\#params} & 0.27M & 0.46M & 0.66M & 0.85M & 1.70M \\
            \hline
            \hline
            ReLU        & 92.52 & 93.21 & 93.78 & 93.85 & 94.41\\
            PReLU       & 92.85 & 93.41 & \textbf{94.12} & \textbf{94.17} & \textbf{94.67}\\
            FALU        & 92.21 & \textbf{93.47} & 92.83 & 93.22 & 92.05\\
            \hline
            GELU        & 92.95 & \textbf{93.47} & 93.08 & 93.12 & \underline{92.99}\\
            FGELU N=1   & \underline{\textbf{93.14}} & 93.39 & \underline{93.22} & \underline{93.54} & 92.83\\
            \hline
            Sig         & 83.56 & 78.27 & 75.56 & 79.58 & 15.42\\
            FSig N=2    & \underline{91.78} & \underline{88.98} & \underline{86.99} & \underline{86.49} & \underline{17.00} \\
            \hline
            Mish        & \underline{92.57} & 92.67 & \underline{92.67} & \underline{92.51} & \underline{92.37}\\
            FMish N=2   & 91.75 & \underline{92.98} & 92.19 & 91.39 & 91.27\\
            \hline
            \hline
        \end{tabular}}
        \label{table:act_fc_cifar10}
    \end{table}
    
    Table~\ref{table:act_fc_cifar10} displays the best test accuracies for various ResNet and activation functions on CIFAR-10.

    \textbf{FSig} accuracies clearly show better performance across all ResNets. We believe that while the sigmoid is not suitable for being an activation function inside networks, changing its shape through fractional calculus helps mitigate some undesirable properties.
    \textbf{FGELU} outperformed GELU in 3 runs. Interestingly, FGELU $N=1$ is identical to GELU and yet it produced a different result. We attribute this fact to the difference in the calculation.
    \textbf{FMish} performed the worst out of our 3 fractional activation functions, being as much as 1.12\% behind in accuracy.
    \textbf{FALU}\footnote{In all experiments, we initialized and kept $a \in [0, 2]$ and $\beta \in [ 1, 10]$.} did not outperform ReLU, PReLU, and GELU in general. Our FALU experimental results differ from the published results (ResNet-18a is architecture-wise ResNet-14 and parameter-wise ResNet-20). The main difference between published results and ours is the higher accuracy of the baseline, non-fractional activation functions.
   
\begin{figure}
        \centering
        \includegraphics[width=.95\linewidth]{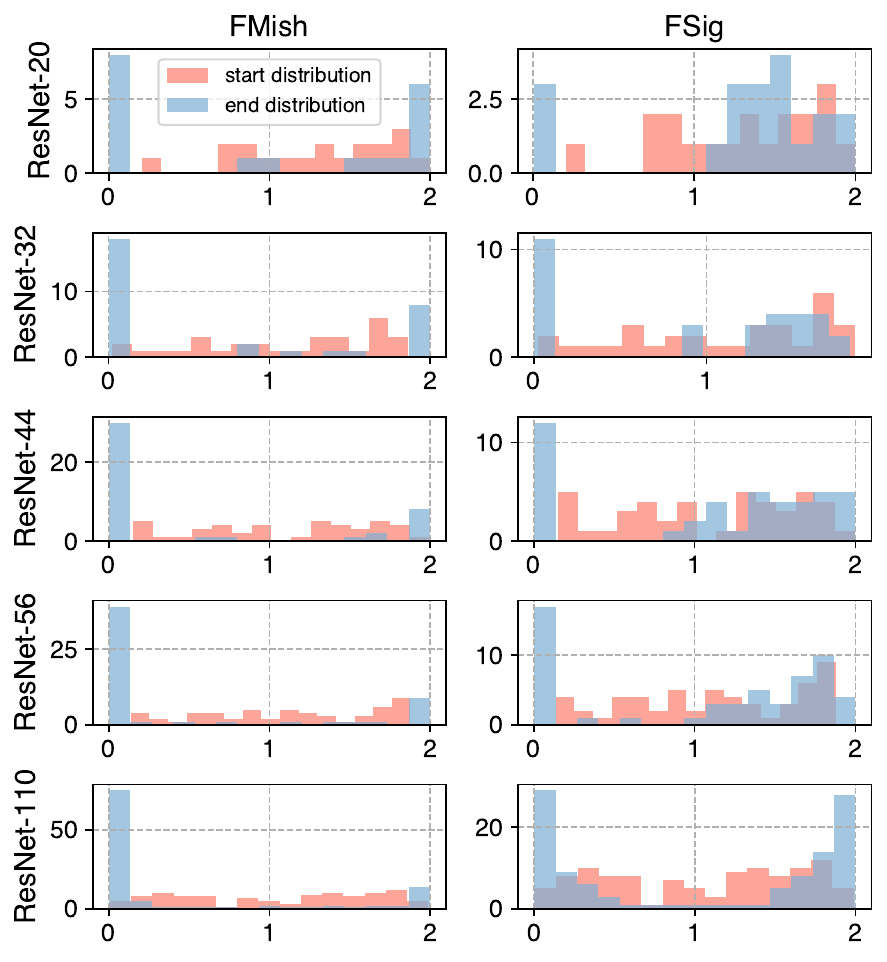}
        \caption{Distribution of FDO at the beginning and the end of the training.}
        \label{fig:resnet_dist}
    \end{figure}

    Figure~\ref{fig:resnet_dist} histograms illustrate the FDO distribution at the beginning and end of the training process across experiments from Table~\ref{table:act_fc_cifar10}. The fractional order tends to converge towards 0 and 2 in all experiments. FMish fractional order progressively moves to 0 as the ResNet depth increases. FGELU is not present in the figure because its fractional order does not affect Eq.~\ref{eq:frac_gelu} when $N=1$.

    Intuitively, fractional activation functions should have an edge over the non-fractional counterparts. However, based on our experiments, this is a case only for sigmoid. The table shows that only two activation functions lead to a consistent increase in performance: ReLU and PReLU, which are the original ResNet paper activation function and its variation.
    
\begin{figure}
        \centering
        \includegraphics[width=\linewidth]{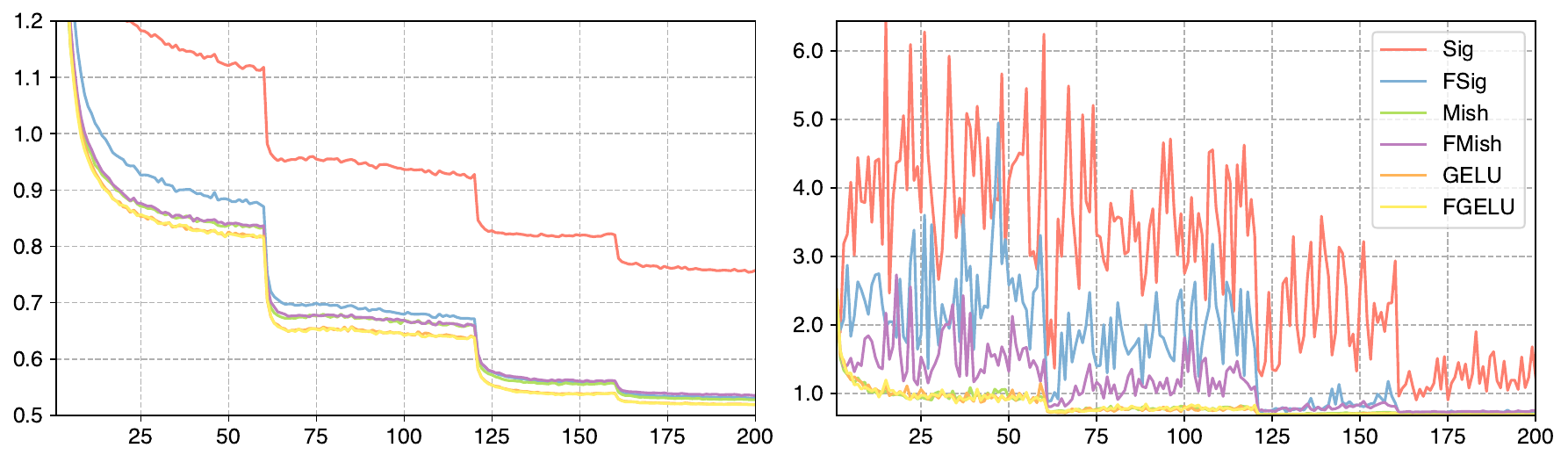}
        \caption{Train (left) and test (right) loss over the training of ResNet-20 on CIFAR-10.}
        \label{fig:resnet_train_test_loss}
    \end{figure}

    We hypothesize that fractional activation functions negatively change the surface of the error function. This is likely due to the complexity of the computation, especially the gamma function. Our thesis is supported by the comparison of train and test loss in Fig.~\ref{fig:resnet_train_test_loss}. While train losses of both fractional and non-fractional activations converge similarly (except for FSig), the test losses do not. Given that the experiments are seeded and deterministic, each model was presented with the same sequence of batches.

    The test loss is computed using test data that lie in proximity to the train data. The significant difference between train and test losses indicates narrow local minima on the loss surface, where any sample-wise divergence leads to a spike in loss, indicating a non-convex surface. This concept is visualized in Fig. 1~\cite{li2018visualizing} that shows how skip connections change the loss surface of ResNets. The fractional activation functions change the ResNet architecture and loss surface. It becomes challenging to find good local minima where the model generalizes well.

    \subsection{EfficientNets}
    We train EfficientNet-B0~\cite{tan2019efficientnet} using three datasets: ImageNet-1K~\cite{ILSVRC15}, CalTech-256~\cite{griffin2007caltech}, and Food-101~\cite{bossard14}. These datasets have in common a higher number of classes, up to 1000, and higher (variable) image resolution. 
    In the case of CalTech-256, we use the train-test split procedure from~\cite{ge2019weakly}, selecting 60 train images per class.

    \textbf{Setting:} We train for 60 epochs with a 2-epoch warm-up. We use SGD with 0.9 momentum, 1e-4 weight decay, and an initial learning rate of 2.5e-2 with a cosine annealing warm restarts scheduler ($T_0=4$, $mult=2$). Data are fed to the network in batches of 64 images augmented with random resized cropping, horizontal flipping, color jitter, and normalization. Training images have a resolution of 224$\times$224, and testing images have a resolution of 320$\times$320~\cite{touvron2019fixing}. Furthermore, we use label smoothing 0.1, gradient clipping by max norm 10, and 16-bit floating precision, identical to sec.~\ref{subsec:resnets}. We track and report the best overall test accuracy. The fractional activation functions have $N$ and $h$ set according to sec.~\ref{sec:hp_tuning} results.
    We change the number of epochs for CalTech-256 to 256 and Food-101 to 90.

    \begin{table}
        \caption{EfficientNet-B0 test accuracies of different dataset/activation function combinations.}
        {\begin{tabular}{l | c c c c }
            \textit{Dataset} & ImageNet-1K & CalTech-256 & Food-101 \\
            \hline
            \hline
            ReLU        & 66.66 & 62.86 & \textbf{85.82} \\
            PReLU       & \textbf{73.41} & 60.12 & 84.10 \\
            SiLU        & 69.09 & 61.39 & 85.12 \\
            FALU        & 67.62 & 59.32 & 83.72 \\
            \hline
            GELU        & 70.78 & 62.62 & \textbf{\underline{85.82}} \\
            FGELU N=1   & \underline{70.99} & \underline{63.13} & 85.77 \\
            \hline
            Sig         & 28.50 & 47.74 & 55.69 \\
            FSig N=2    & \underline{60.18} & \textbf{\underline{63.17}} & \underline{83.82} \\
            \hline
            Mish        & \underline{54.82}$^*$ & \underline{60.97} & \underline{85.35} \\
            FMish N=2   & 53.69$^*$ & 58.53 & 00.00$^\ddag$ \\
            \hline
            \hline
        \end{tabular}}
        
        \label{table:act_fc_effnetb0}
    \end{table}
    
    Table~\ref{table:act_fc_effnetb0} shows similar results to Table~\ref{table:act_fc_cifar10}. The performance difference in accuracy of the fractional and non-fractional activation functions holds across the datasets. A surprising result is the performance of PReLU on ImageNet-1K that is significantly higher than SiLU (the default EfficientNet activation function).

    Using FMish as EfficientNet-B0 activation leads to crashing of the training process on ImageNet-1K. In order to run full training, we used stricter gradient clipping with a maximum norm of 1 ($^*$ result in Table~\ref{table:act_fc_effnetb0}). The stricter gradient clipping also led to fewer oscillations and spikes in test loss (Fig.~\ref{fig:effnet_train_test_loss}). Gradient clipping failed to stabilize training of FMish on Food-101 ($^\ddag$ result in Table~\ref{table:act_fc_effnetb0}).

    \begin{figure}
        \centering
        \includegraphics[width=.95\linewidth]{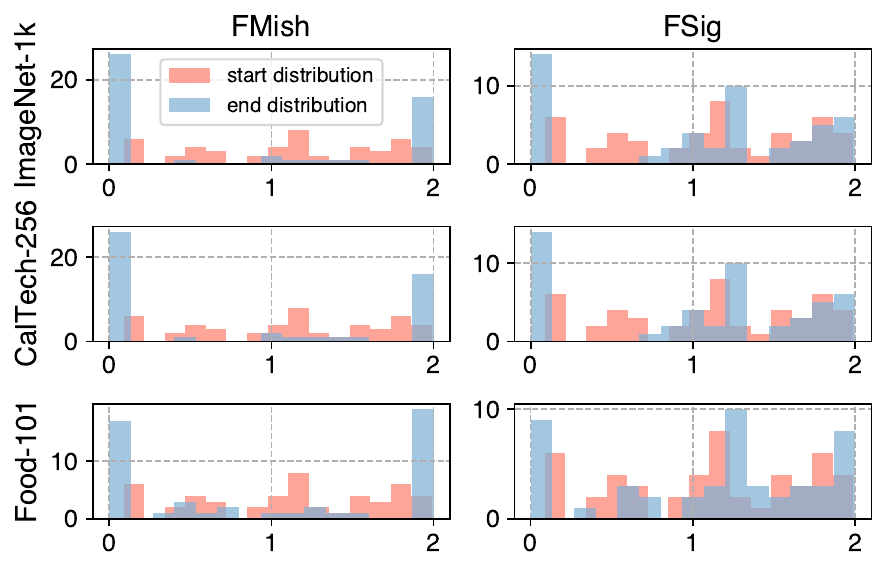}
        \caption{Distribution of FDO at the beginning and the end of the training.}
        \label{fig:effnet_dist}
    \end{figure}

    Figure~\ref{fig:effnet_dist} histograms show mostly similar behavior to ResNet experiments. FSig tends to accumulate more around the first derivative. FMish derivative distributions are accumulating around zero and the second derivative.

    \begin{figure}
        \centering
        \includegraphics[width=\linewidth]{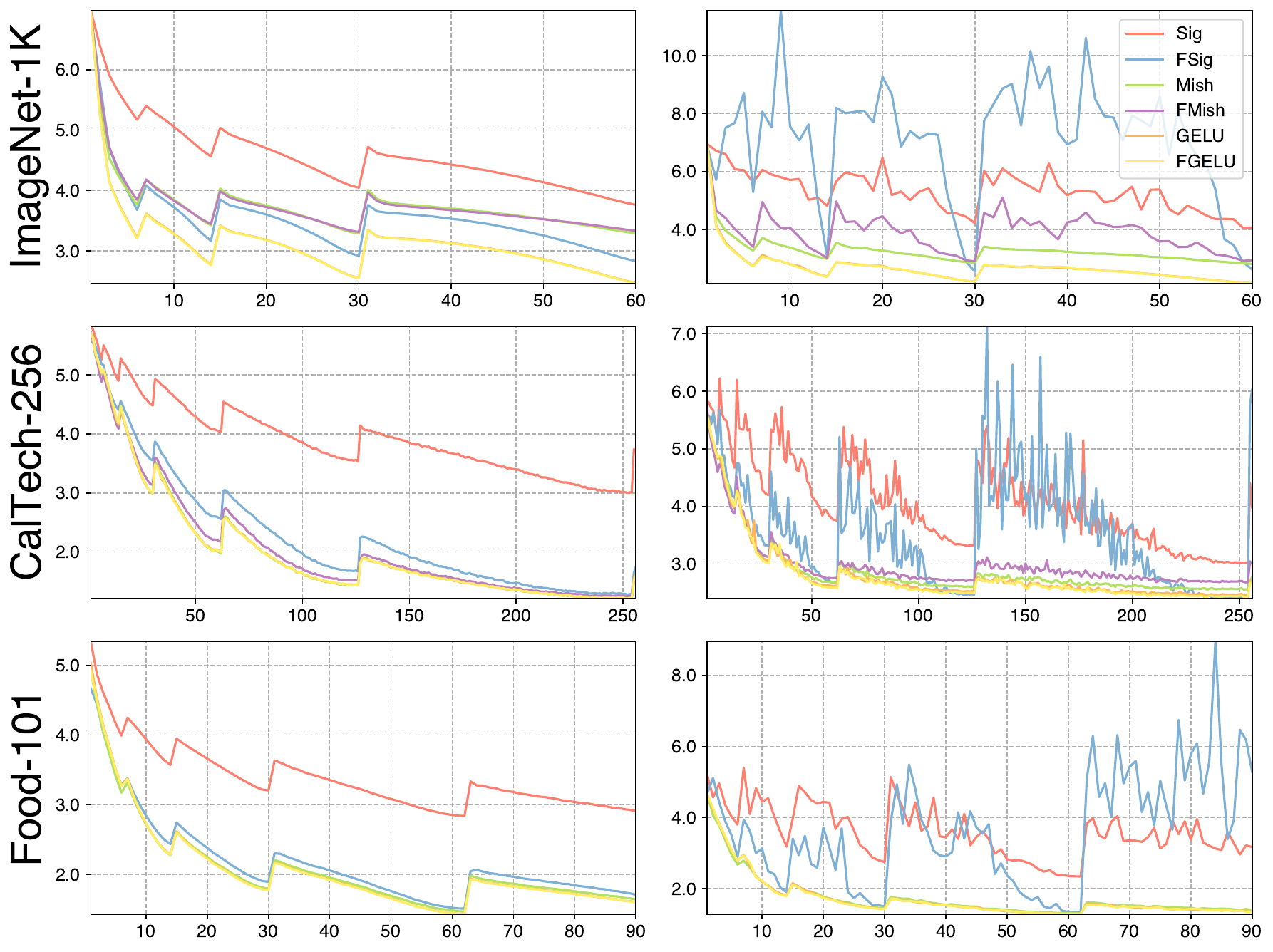}
        \caption{Train (left) and test (right) loss evolution over the training of EfficientNet-B0.}
        \label{fig:effnet_train_test_loss}
    \end{figure}

    Training and testing losses generally follow the same pattern as the losses in sec.~\ref{subsec:resnets}. The spikes in training and testing losses are caused by learning rate restarts.

\section{Conclusion}
    \label{sec:conclusion}
    Fractional activation functions have emerged as a promising alternative to traditional activation functions. By introducing a FDO as a tunable parameter, these functions offer a more flexible and expressive representation of activation dynamics. This paper has explored fractional activation functions, detailing their theoretical foundations and practical implementation challenges. We have reviewed existing work and introduced new functions FGELU and FMish.
    
    Our experimental evaluation of various fractional activation functions revealed that, in certain cases, they can outperform their non-fractional counterparts. Notably, the fractional Sigmoid function demonstrated improved performance across several experiments. However, the overall consistency of fractional activation functions remains less reliable compared to traditional activation functions.
    
    The ablation study highlights that the time and computational complexity of fractional activation functions do not scale favorably with increasing parameter \( \Sigma \). Addressing this issue is crucial for future research. Specifically, future work should focus on optimizing complexity scaling and examining the impact of different \( \Sigma \) values on performance.
    
    Despite these observed inconsistencies, we are optimistic about the potential of fractional activation functions to enhance neural network performance. We advocate for continued research to better understand and leverage these functions in practical applications.
    
    Our code and experimental details are publicly available at \url{https://gitlab.com/irafm-ai/frac_calc_ann}.

\section*{Acknowledgment}
  The study described is from the project Research of Excellence on Digital Technologies and Wellbeing CZ.02.01.01/00/22\_008/0004583 which is co-financed by the European Union.

  This article has been produced with the financial support of the European Union under the REFRESH – Research Excellence For REgion Sustainability and High-tech Industries project number CZ.10.03.01/00/22\_003/0000048 via the Operational Programme Just Transition.

\bibliographystyle{plain}
\bibliography{refs}

\end{document}